\documentclass[journal]{IEEEtran}

\usepackage{mathtools}

\usepackage{float}
\usepackage[colorlinks,linkcolor=black,anchorcolor=black,citecolor=black]{hyperref}
\usepackage[pdftex]{graphicx}
\graphicspath{{./graphics/}}
\usepackage{subfigure}
\usepackage{multicol}
\usepackage{paralist}
\usepackage{bm}
\usepackage{amsfonts}
\usepackage{url}
\usepackage{multirow}
\usepackage{textcomp}
\usepackage{color}
\usepackage{diagbox}
\usepackage{graphics}
\usepackage{booktabs}
\usepackage{cite}

\makeatletter
\newif\if@restonecol
\makeatother

\usepackage[linesnumbered,ruled,vlined]{algorithm2e}
\usepackage{algpseudocode}
\usepackage{amsmath}

\begin{document}
	\pagestyle{empty}
\title{OGNet: Salient Object Detection with Output-guided Attention Module}

\author{{Shiping Zhu, \IEEEmembership{Member, IEEE}, Lanyun Zhu}

\thanks {This work is supported by the National Natural Science Foundation of China (NSFC) under grant No. 61375025, No. 61075011 and No. 60675018, and also the Scientific Research Foundation for the Returned Overseas Chinese Scholars from the State Education Ministry of China.}
\thanks {Shiping Zhu and Lanyun Zhu are with the Department of Measurement Control and Information Technology, School of Instrumentation Science and Optoelectronics Engineering, Beihang University, 100191, HaiDian District, XueYuan Road No. 37., Beijing, China. (phone: +86-13391687912; e-mail: spzhu@163.com)}}

\maketitle
\thispagestyle{empty}

\begin{abstract}
  Attention mechanisms are widely used in salient object detection models based on deep learning, which can effectively promote the extraction and utilization of useful information by neural networks. However, most of the existing attention modules used in salient object detection are input with the processed feature map itself, which easily leads to the problem of `blind overconfidence'. In this paper, instead of applying the widely used self-attention module, we present an output-guided attention module built with multiscale outputs to overcome the problem of `blind overconfidence'. We also construct a new loss function, the intractable area F-measure loss function, which is based on the F-measure of the hard-to-handle area to improve the detection effect of the model in the edge areas and confusing areas of an image. Extensive experiments and abundant ablation studies are conducted to evaluate the effect of our methods and to explore the most suitable structure for the model. Tests on several datasets show that our model performs very well, even though it is very lightweight.

\end{abstract}


\IEEEpeerreviewmaketitle

\begin{figure*}[t]
	\centering
	\includegraphics[width=1\linewidth]{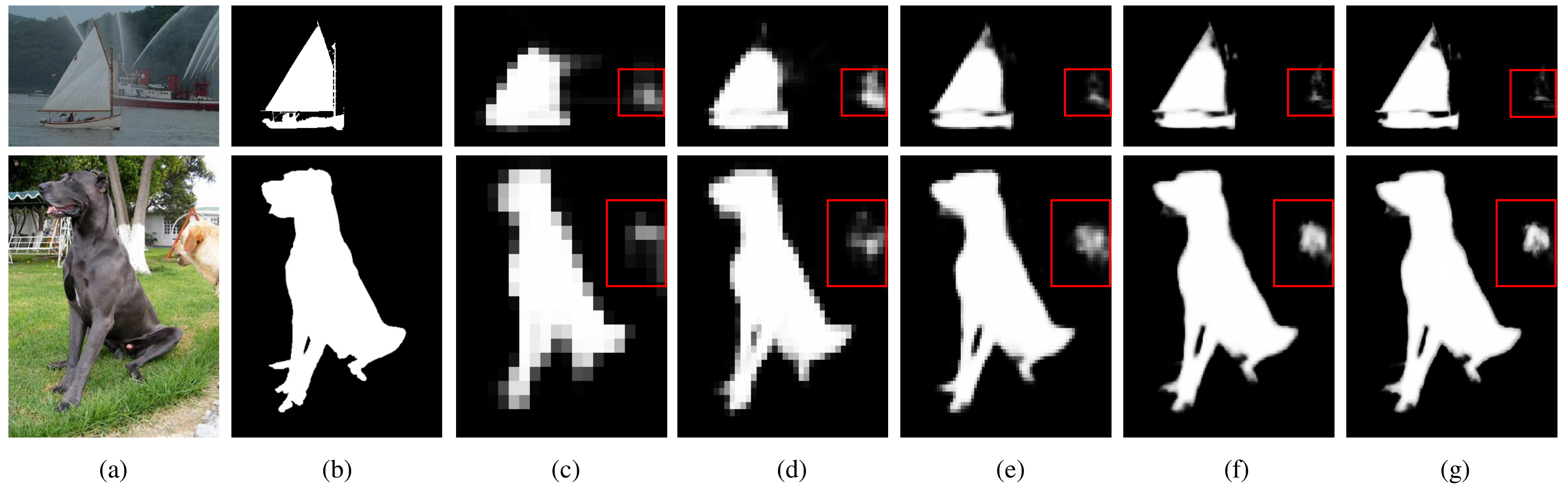}
	\caption{Some examples of output in different layers. (a) Input image; (b) ground truth; (c) output of layer 1; (d) output of layer 2; (e) output of layer 3; (f) output of layer 4 (g) output of layer 5. The area in the red box is where the output is misjudged. For the first line, output of the shallower layers makes fewer mistakes. For the second layers, output of the shallower layers makes more mistakes.}
	\label{1}
\end{figure*}
\section{Introduction}
\IEEEPARstart{S}{alient} object detection aims to estimate the region of the most attractive object in an image, and it is an important research area in computer vision. It has great application value in the fields of scene classification\cite{1}, object detction\cite{67}, image retrieval\cite{2,3} and visual tracking\cite{50,51,60}. Salient object detection is a very challenging problem because it requires both a correct identification of the salient object and an accurate display of the salient region. In recent decades, many algorithms for salient object detection have been proposed. Inspired by the human visual attention mechanism, traditional unsupervised methods\cite{4,5,6,7,8,68} typically apply handcrafted features in images to determine the salient region. These methods do not perform well when the background of the image or the shape of the salient object is very complicated.\par
Recently, deep learning has made rapid development, and many methods based on deep learning\cite{18,21,23,26,28,44,46,47,49} have greatly improved the accuracy of salient object detection. Models based on convolutional neural networks and recurrent neural networks have achieved remarkable performance in many tasks, such as image classification\cite{9,10}, object detection\cite{11,12,13} and semantic segmentation\cite{14,15}. A deep neural network can effectively extract and fuse different levels of features in images, which effectively solves the insufficiency of image feature extraction and fusion in traditional methods. Many salient object detection models based on deep learning adopt the encoder-decoder as the basic structure of the neural network\cite{35,36,47}. This structure, represented by FCN\cite{16}, reduces the image resolution by passing the encoder and extracting the image features from different levels; and then, it gradually recovers the image resolution by the decoder and finally gains the saliency map. The encoder-decoder structure is widely adopted since it can recover the contour shape of the salient objects well.\par
Since the encoder-decoder has a weaker extraction ability for semantic information, the simple utilization of the encoder-decoder cannot gain quite an outstanding performance. Hence, many research studies are committed to improving the original encoder-decoder structure. Two methods can upgrade the detection effects by increasing only a small memory footprint and the cost of computing. One is to use networks with multiscale outputs\cite{17}. Different from most models that have only one output, the multiscale output structure, represented by deeply supervised net,  obtains many outputs in various positions of the neural network. Such a structure can make the deeper parts of the neural network easier to train and lead the network to place more emphasis on the required tasks, avoiding information turbulence and mistakes. Many models based on multiscale output structures for image segmentation\cite{20}, object detection\cite{19} and salient object detection\cite{17,18,21,58} have achieved good performance. The other method uses the attention mechanism. In recent years, the attention mechanism has become one of the most important research directions of deep learning\cite{22} because it can significantly improve the effect of models by adding only a small amount of computation. Attention mechanisms can reinforce useful or key information and impair useless or incorrect information. Salient object detection is a task of classifying each pixel into two categories, and the introduction of an attention mechanism can enhance the confidence for the model's judgement.\par
In this paper, we make full use of the features of these two structures. Traditional attention modules used in segmentation and salient object detection tasks usually apply the processed feature maps themselves as the input of the attention modules. In such a structure, it is easy to realize two kinds of favorable operations, reinforcing `true positive' information and weakening `false negative' information, as well as generate two kinds of faulty operations, reinforcing `false positive' information and weakening `true negative' information. This is a problem we call the `blind overconfidence' of the attention module judgement. To solve this problem, we take the deeper layer's low-resolution output as the input of the shallower layer's attention module to establish a new output-guided attention module. Compared with the ordinary self-attention module, taking the deeper layers' output as the input of the attention module can integrate the advantage of each layer of the neural network, thus preventing the attention module of decoder in one layer from enlarging the false information caused by this layer. Considering that different input feature maps have different importance in attention module processing for different input images, in order to reasonably select multiple input of  the attention module, we make the network learn a set of weights. And all the input features maps are weighted before fed into the attention module. Based on the output-guided module, we built a new salient object detection model applying the classic encoder-decoder structure. Our model gains remarkable detection effects with a small memory footprint and fast detection speed. Moreover, with the enlightenment of the features of outputs in different layers, we propose a method to identify regions that are difficult to estimate in images of the training set. On this basis, we propose the intractable area F-measure loss function, which can pay more attention to the areas that are difficult to judge in the image. The main contributions of this paper can be summarized as follows:
\begin{itemize}
	\item We propose a new output-guided attention module built with outputs in various positions of an neural network, which can overcome the shortcomings of many other self attention modules. 
	\item  We propose a new end-to-end neural network for salient object detection applying the output-guided attention module.
	\item  We propose an intractable area loss function based on the features of the multi-output structure. The introduction of this loss function makes the model more effective facing complicated images.
\end{itemize}
\par
The rest of this paper is organized as follows: in Section \ref{relatedwork}, we make a conclusion about the existing classic salient object detection models and attention modules; in Section \ref{method}, we introduce the output-guided attention module, OGNet and intractable area F-measure loss; in Section \ref{exp}, we demonstrate our experimental results during the research process. Finally, we conclude this paper in Section \ref{conclusion}. 

\section{Related Work}\label{relatedwork}
\subsection{Salient Object Detection}
 In the early stages of development, salient object detection models were usually based on low-level hand-crafted features, such as color features\cite{4} and textural features\cite{7,22,66}. Although these models generated certain effects, their performances were not ideal for images with complicated backgrounds or complex salient objects. When making saliency judgments, the human eyes always confront complicated elements, while in traditional methods, fully considering and integrating various factors are difficult. Deep learning explores a new route for the research of salient object detection. Early salient object detection models based on deep learning usually select the convolutional layer - fully connected layer structure, which is the same as most image classification models. Wang {\emph {et al}}.\cite{23} proposed two neural networks to detect salient objects, one for learning local patch features to determine the saliency value of each pixel and the other for predicting the saliency score of each object region based on global features. Li and Yu\cite{24} first segmented the image into several areas and then formulated a neural network with some branches to train these areas. Then, their method utilizes several convolutional layers to connect them together to achieve information integration among different layers. Zhao {\emph {et al}}.\cite{25} built a multicontext deep learning framework with two branches that extract global context and local context and then integrate them together. After the appearance of fully convolutional networks, many salient object detection models based on deep learning adopted the encoder-decoder structure represented by FCN and then made some adjustments to that structure. Liu and Han\cite{18} made use of the hierarchical recurrent convolution to build up the decoder part of the neural network. Zhang {\emph {et al}}.\cite{26} applied the reformulated dropout to some convolutional layers on the strength of the basic encoder-decoder structure to extract the salient information more conveniently. However, due to the inadequate use of different levels of information, it is difficult to achieve very good performance in a simple encoder-decoder. Hou {\emph {et al}}.\cite{17} utilized a large number of short connections to join the decoders in different layers together and drew on the idea of DenseNet\cite{27}, which worked to realize the full integration of information in different layers. Similarly, Zhang {\emph {et al}}.\cite{28} proposed the Amulet. Wang {\emph {et al}}.\cite{29} proposed a multistage structure and used pyramid pooling in the joint part to obtain and merge the information from different layers together. Such methods usually perform well. However, due to a demand to connect feature maps in different layers, they often need to consume a large amount of memory and require a large amount of computation. 
 
 \subsection{Attention Mechansim}
 During deep learning, the attention mechanism was applied to the field of machine translation\cite{30,31} at the earliest stages and accomplished outstanding effects. Then, it is applied to the neural network models of computer vision. For the past few years, many models applying attention mechanisms have greatly improved the effects in image classification\cite{32}, semantic segmentation\cite{33,54}, action recognition\cite{61} and other fields. The core ideology of the attention mechanism is to selectively enhance or weaken the large amount of information constructed by neural networks. The attention module of SENet proposed by Hu {\emph {et al}}.\cite{32} includes two processes: squeeze and excitation. The squeeze process applies global average pooling to compress the feature maps, and the excitation process utilizes two fully connected layers to obtain a series of weights, which are used to weigh feature maps from channels. This method improves the accuracy of image classification models immensely. Furthermore, CBAM proposed by Woo {\emph {et al}}.\cite{33} expands the attention mechanism’s treatment dimension from the channel dimension of SENet to two dimensions — channel and spatial — and selects both the average value and maximum value to compress the feature maps, which further increases the effects of the attention module. The structures of SENet and CBAM can expand to many other computer vision task models. In recent years, many salient object detection models have also utilized various kinds of attention modules. The module proposed by Zhang {\emph {et al}}. builds two attention modules from the channel and spatial layers, which is similar to the establishment method of CBAM.  Liu {\emph {et al}}.\cite{36} applied a convolution and bidirectional LSTM to formulate local pixelwise attention and global pixelwise attention , which enlarges the receptive field to reduce mistakes. All the attention modules mentioned above only use the processed feature maps themselves as the input, which is the main difference between the proposed method and other methds for salient object detection only using self attention \cite{36,35}.
\section{Methodology}\label{method}
In this section, we introduce our proposed methods. The output-guided attention module is introduced in Subsection \ref{attention} and the complete structure of the output-guided network (OGNet) is shown in Subsection \ref{OGnet}. The intractable area F-measure loss and the training method are shown in Subsection \ref{IAFloss} and Subsection \ref{train}, respectively.
  \begin{figure}
 	\centering
 	\includegraphics[width=1\linewidth]{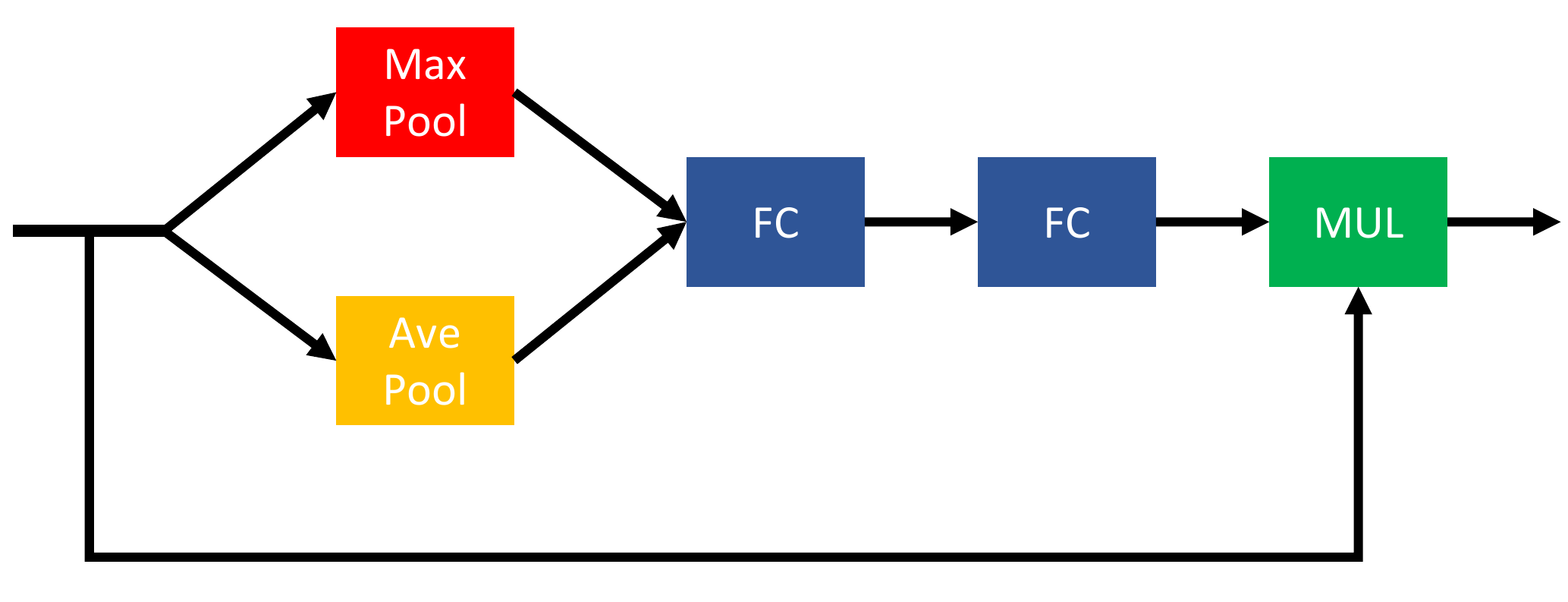}
 	\caption{The structure of channel attention. FC is the fully connected layer. MUL is a multiplication operation.}
 	
 	\label{catt}
 \end{figure}

\begin{figure}
	\centering
	\includegraphics[width=1\linewidth]{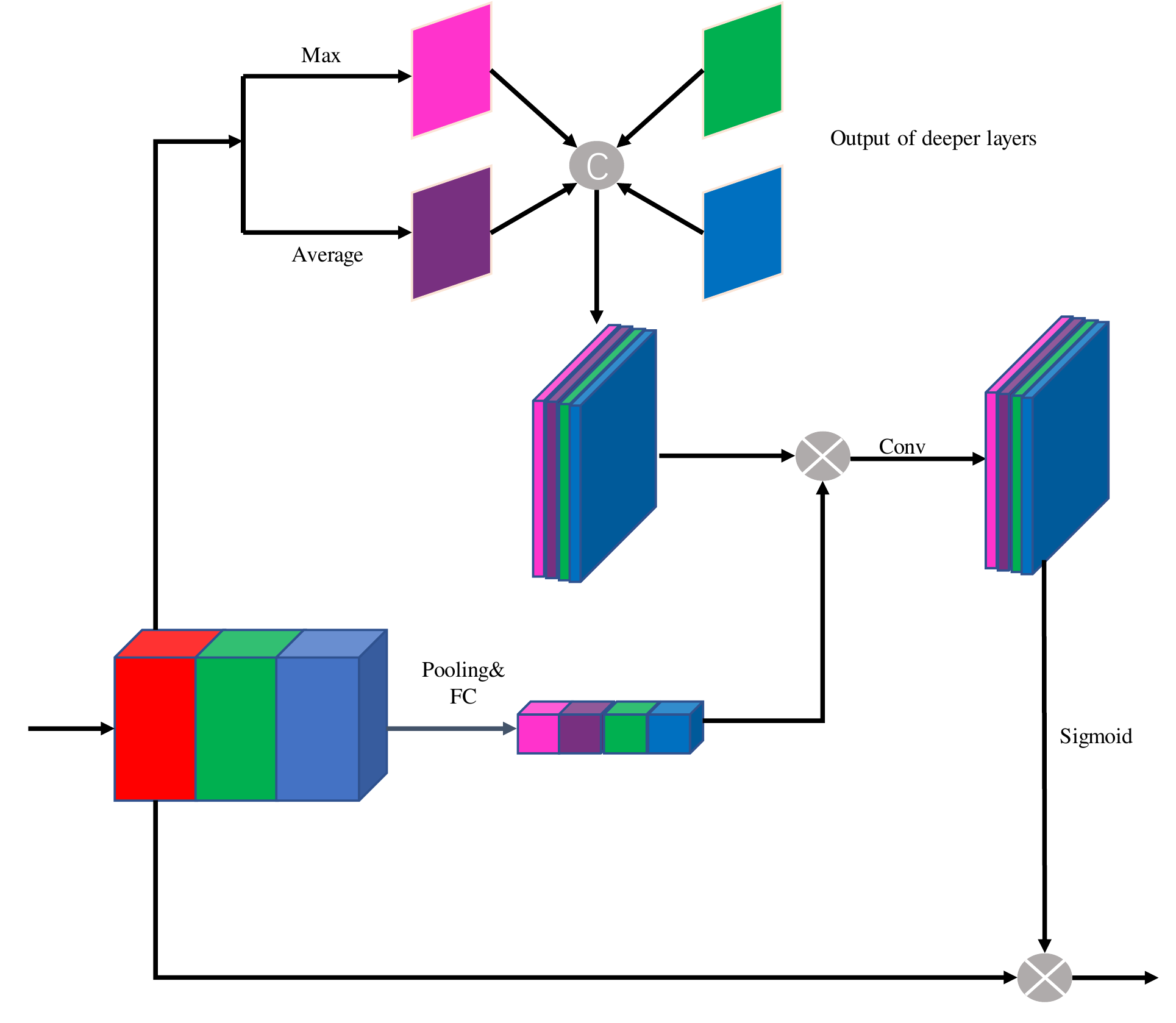}
	\caption{The structure of spatial attention. CAT is the concatenation of some groups of feature maps from channels.}
	\label{satt}
\end{figure}

\begin{figure*}
	\centering
	\includegraphics[width=1\linewidth]{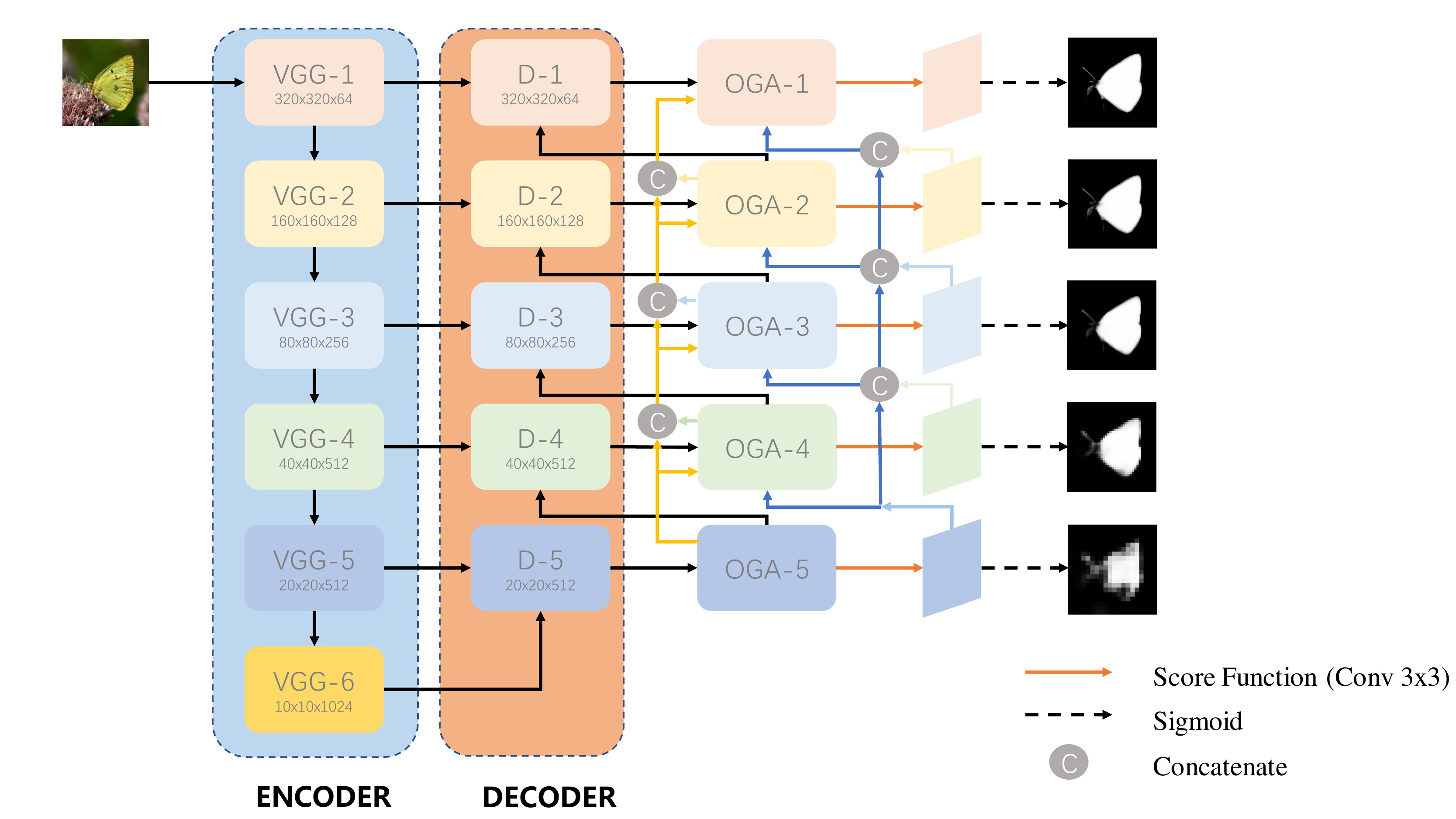}
	\caption{The detailed structure of the output-guided model.}
	\label{OG}
\end{figure*}
\subsection{Output-guided Attention Module}\label{attention}
\subsubsection{Blind Overconfidence}\label{blind}
At present, attention mechanisms composed of both spatial attention and channel attention, represented by CBAM\cite{34}, is one of the most popular attention modules used in various kinds of computer vision models. CBAM builds up two attention modules - channel attention and spatial attention, taking the processed feature maps themselves as input. The effects of models can be greatly improved through these two attention modules. Such a module is very suitable for image classification because image classification does not concern the shape and location of objects in an image; thus, the enhancement of incorrect information caused by attention modules will not have a great impact on the final judgment. However, we think that this kind of spatial attention, which only takes processed feature maps as input, faces some problems when applied to salient object detection. Salient object detection aims to classify each pixel in an image into two categories, which means that the final saliency map is binary. Assuming that a pixel in an image is salient, the value of the corresponding position in the feature maps should be large. If a layer of the neural networks misjudges the pixel, the following attention module will greatly magnify the wrong information, which is the problem of `blind overconfidence'  in attention modules. An ideal attention module should magnify the correct information and avoid the enhancement of wrong information. A good resolution is to enlarge the receptive field of the attention module to capture more information. However, this requires a lot of computation. To solve the problem of ‘blind overconfidence’, we built a new attention module. Similar to CBAM, it consists of both channel attention and spatial attention. The structures of channel attention module and spatial attention module are shown in Fig. \ref{catt} and Fig. \ref{satt}, respectively.
 \subsubsection{Channel Attention}\label{channel}
  In a layer of neural networks, not all feature maps have the same significance. The channel attention module is a feature detector that can enhance information in useful feature maps and reduce information in useless feature maps. If we adopt the whole feature map as the input of the attention module, the computation will be quite large, which violates the design principle that the attention module should be lightweight. Thus, we should find a method whose receptive field is large enough to express the global feature of a feature map. Similar to CBAM, we use both max pooling and average pooling to demonstrate the global feature of the input feature map ${\mathbf F}\in {\mathbb R}^{C \times W \times H}$. The channel attention map can be calculated as follows:
  \begin{equation}
  {\mathbf W}_{\mathbf c}=Sigmoid(L_2(L_1(GMP({\mathbf F})))+L_2(L_1(GAP({\mathbf F}))))
  \end{equation}
  where {\emph GMP} is global max pooling and {\emph GAP} is global average pooling. The input size of \emph{L$_{1}$} and the output size of \emph{L$_{2}$} are \emph{C}. The input size of \emph{L$_{2}$} and the output size of \emph{L$_{1}$} are \emph{C/4}.
  This setup is designed to deepen the network to extract more information with reducing the additional memory footprints caused by these two fully connected layers. \emph{L$_{1}$} is followed by a rectified linear unit (ReLU)\cite{52}. Note that \emph{L$_{1}$} and \emph{L$_{2}$} are shared for feature maps after max pooling and average pooling. \emph{Sigmoid} is a function used to get the attention map. For each resolution $\mathbf{X}_\mathbf{c}^i$ in a feature map processed by two fully networks:
  \begin{equation}
  \mathbf{W}_\mathbf{c}^i=\frac{1}{1+e^{-\mathbf{X}_\mathbf{c}^i}}
  \end{equation}
\subsubsection{Spatial Attention}
Spatial attention is used to enhance the confidence of the model on its judgement. In salient object detection, the use of spatial attention can also make a model focus on the foreground region, which is beneficial for saliency prediction. Unlike CBAM, apart from taking the average and maximum value from channels, outputs from other layers are also taken as the input. We think that taking the outputs of other layers into the attention module is a kind of balance and compensation, which can avoid the enhancement of wrong information caused by the attention module in one layer. Beyond that, this structure can also be regarded as a special form of short connection, which can make full use of information from different layers and make the deep neural network easier to train. For feature maps from the decoder in layer \emph{m}, we obtain two feature maps, $\mathbf{F}^{max}_m$ and $\mathbf{F}^{min}_m$, which express the comprehensive information of all layers by calculating the maximum and average values on the channel. For layer \emph{m}, the input of the attention module is $\{\mathbf{F}^{max}_m, \mathbf{F}^{min}_m,\mathbf{O}_{m+1},...,\mathbf{O}_M\}$, where $\mathbf{O}_{i}$ is the output of the i$^{th}$ layer.\par
A straightforward idea is to input these maps directly into some convolution layers to obtain spatial attention weight. However, this approach regards all maps as having the same importance and ignores the differences between them. As shown in Fig. \ref{1}, for the first column, output maps in shallower layers make fewer errors in saliency judgments for the area in the red box. Thus, when fed into the attention module, shallower output should be more important. However, for the second column where outputs in deeper layers judge better, deeper output should be emphasized. Thus, weighting these maps before feeding them into the spatial attention module is necessary. The weight is also obtained from the neural network. We first concatenate $\mathbf{F}_m$ with output of output-guided attention modules of all deeper layers $\{\mathbf{OG}_{m-1}, \mathbf{OG}_{m-2},...,\mathbf{OG}_M\}$ to get a feature map $\mathbf{C}$. $\mathbf{C}$ passes through two fully connected layers which are similar to that in channel attention module and obtain a vetor $\mathbf{V}$ with $M-m$ dimensions. Finally, the spatial attention map can be generated as follows:
\begin{equation}
\mathbf{W}_{\mathbf{s}}=Sigmoid(f^{7\times 7}(\mathbf{	V}.CAT(\mathbf{F}_m^{max},\mathbf{F}^m_{min},\mathbf{O}_{m-1},...,\mathbf{	O}_M)))
\end{equation}
where $f^{7\times 7}$ is a $7\times 7$ convolution layer. The size of the convolution layer is bigger than the usual one which is $3\times 3$ because the receptive field should be large enough to fully extract pixel relationship for the sptial attention. \emph{CAT} is the concatenation of feature maps from channels. \par
When utilizing the attention module, we let the processed feature maps pass the channel attention module first and then pass the spatial attention module to obtain the final output of the output-guided attention module. The output can be obtained as follows:
\begin{equation}
{\mathbf F}_{out}={\mathbf W_s}.{\mathbf W_c}.{\mathbf F}_{in}
\end{equation}
 This arrangement that passes the channel attention module first is also inspired by\cite{34}, which argues that channel-first is slightly better than spatial-first. 
\subsection{OGNet}\label{OGnet}
Based on the output-guided attention module introduced above, we propose a new model for the salient object detection: output-guided model (OGNet). Our model strengthens the basic encoder-decoder structure. To compare fairly with most salient object detection models, we choose the most commonly used VGG16\cite{37} as the backbone of the encoder. Note that, similar to most models, the backbone can be flexibly selected and can be replaced by other networks, such as ResNet\cite{38} and Xception\cite{39}. \par
\begin{figure}[t]
	\centering
	\includegraphics[width=1\linewidth]{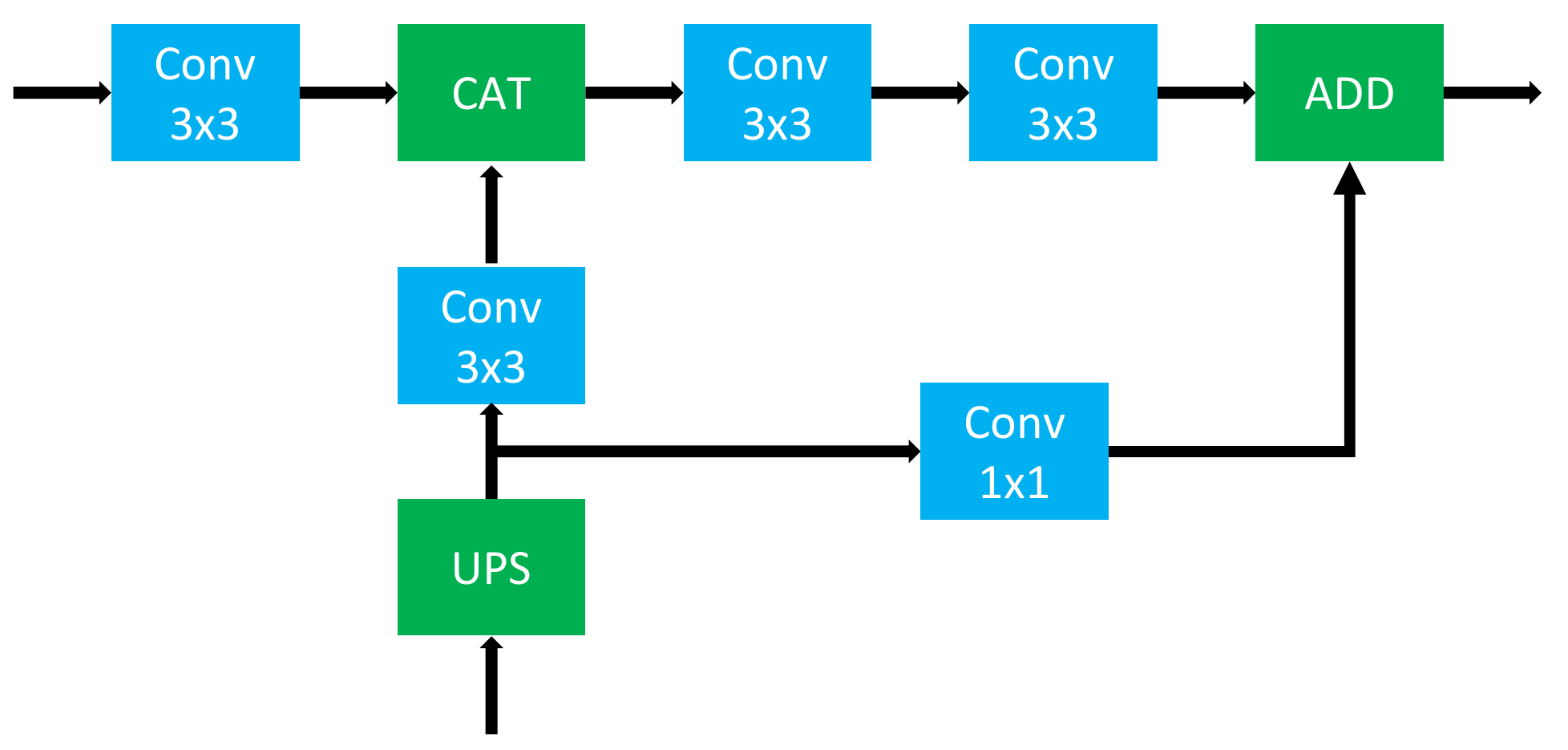}
	\caption{The detailed structure of a layer of the decoder. Feature maps from encoder and decoder are input from the left and bottom, respectively. UPS is the bilinearly upsamping.}
	\label{decoder}
\end{figure}

 
Our model's decoder contains five layers so that five saliency maps with different resolutions are gained. Each layer of the decoder has the same structure. The structure of the decoder is shown in Fig. \ref{decoder} and details of each convolution are shown in Table \ref{decoder_structure}. The ith layer of the decoder takes the output of the encoder in the same layer and the output of the previous layer's decoder as input. First, the decoder feature maps are bilinearly upsampled by a factor of 2, and then two $3\times 3$ convolutions are applied on feature maps from the encoder and decoder separately. Note that we do not use deconvolution directly because bilinearly upsampling performs slightly better than deconvolution. Inspired by\cite{17}, we tried to use a larger-sized convolution such as $7\times 7$ and $5\times 5$ to process feature maps from encoder but found that it could not improve performance but instead caused overfitting. We performed some experiments to find the most suitable convolution size, and the results are shown in Section 4.3. The encoder feature maps and decoder feature maps are concatenated, and another two $3\times 3$ convolutions are applied to further fuse and extract information from the feature maps. Inspired by the structure of ResNet\cite{38}, a residual block is applied to construct the decoder. For each layer of the decoder, we apply a $1\times 1$ convolution to convert the feature map which has been bilinearly upsampled to the same number of channels as the output of the decoder in this layer. Then this feature map is added to the output of the decoder to obtain the final output. Section 4.3 shows the comparison between the performance of models using residual blocks and not using residual blocks.\par
The output of every layer of the decoder passes an output-guided attention module, which is the input of the decoder in the next layer, as well as passes a $3\times 3$  convolution and Sigmoid function to obtain this layer's output saliency map. Note that the inputs of the output-guided attention module are the saliency maps that have not passed the Sigmoid function. All convolutions in the decoder are followed by a batch normalization and ReLU. The structure of the output-guided model is shown in Fig. \ref{OG}.

\begin{table}
    \linespread{1.5}
	\centering 
	\caption{Structure of each layer of the decoder. conv\_e, conv\_d represent the convolution layers processing input from encoder and decoder. conv\_1 and conv\_2 refer to two convolution layers after the concatenation.}
	\label{decoder_structure}
	\begin{tabular}{c|c|c|c|c}
		\toprule
		\diagbox{No.}{Layer}&conv\_e&conv\_d&conv\_1&conv\_2\\
		\midrule
		1&$3\times 3,128$&$3\times 3,128$&$3\times 3,256$&$3\times 3,256$\\
		2&$3\times 3,128$&$3\times 3,128$&$3\times 3,256$&$3\times 3,256$\\
	3&$3\times 3,64$&$3\times 3,64$&$3\times 3,128$&$3\times 3,128$\\
	4&$3\times 3,32$&$3\times 3,32$&$3\times 3,64$&$3\times 3,64$\\
		5&$3\times 3,32$&$3\times 3,32$&$3\times 3,64$&$3\times 3,64$\\
		\bottomrule
	\end{tabular}
\end{table}
\begin{figure*}[t]
	\small
	\centering
	\includegraphics[width=0.8\linewidth]{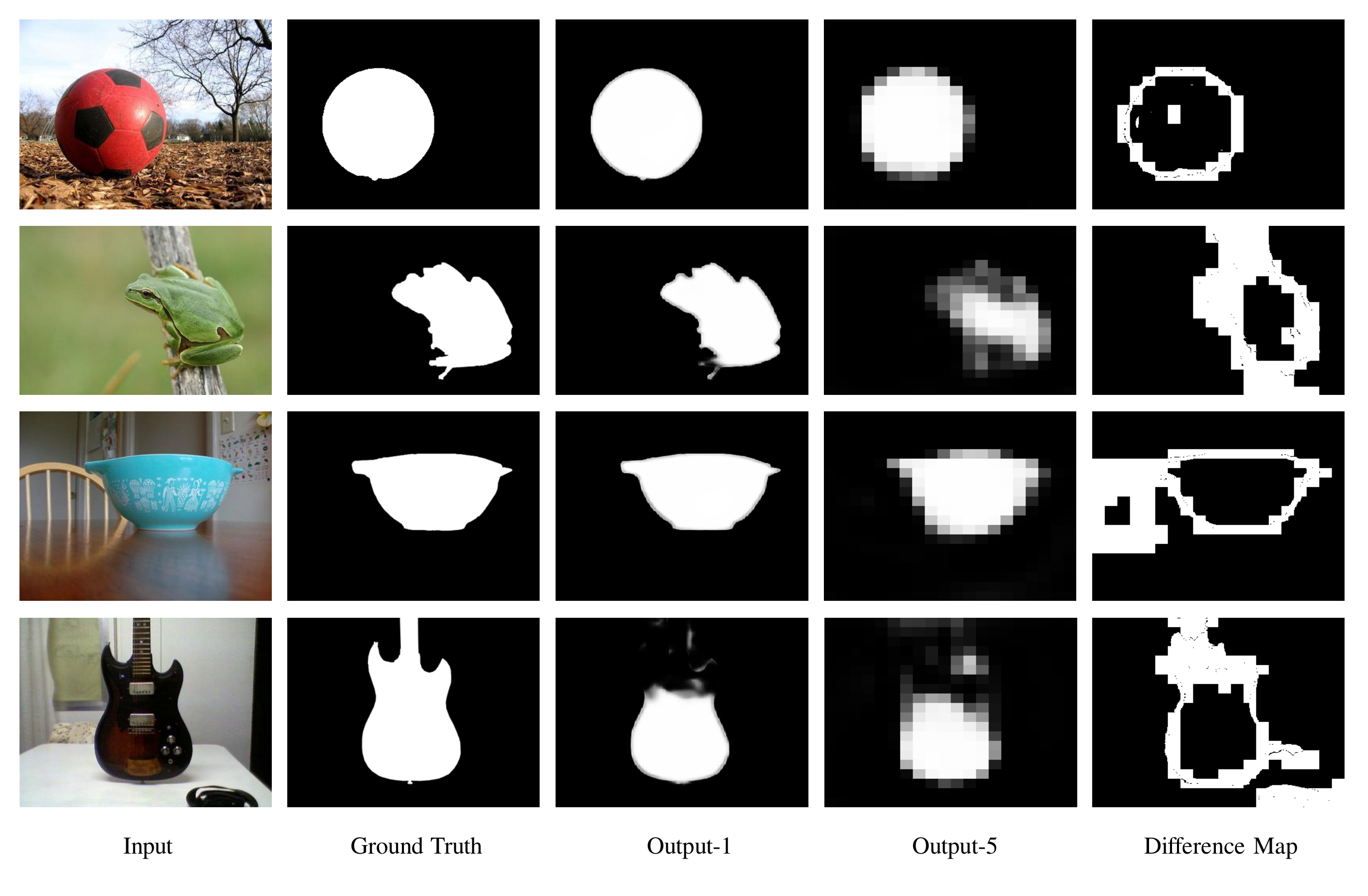}
	\caption{Some examples of multi output and the difference map. The output of different layers in the network is different in some areas. From the difference maps, we can find that the different areas are usually the boundary of the objects or where the disturbing objects are located.}
\end{figure*}
   \begin{table*}
	\linespread{1.5}
	\centering 
	\footnotesize
	\caption{Quantitative comparison of MAE, F-measure and S-measure with 15 methods on 5 datasets. A higher F-measure score, higher S-measure score and lower MAE score represent better performance. The top three results are highlighted in {\color{red}red}, {\color{green}green} and {\color{blue}blue}, respectively.}
	\begin{tabular}{lr|ccc|ccc|ccc|ccc|ccc}
		\toprule
		\multicolumn{2}{c|}{	\multirow{2}*{\diagbox{Methods}{Datasets}}}&\multicolumn{3}{c|}{HKU-IS}&\multicolumn{3}{c|}{ECSSD}&\multicolumn{3}{c|}{SOD}&\multicolumn{3}{c|}{DUT-OMRON}&\multicolumn{3}{c}{DUTS-TE}\\
		\midrule
		~&~&MAE&$\mathcal{F}$&$\mathcal{S}$&MAE&$\mathcal{F}$&$\mathcal{S}$&MAE&$\mathcal{F}$&$\mathcal{S}$&MAE&$\mathcal{F}$&$\mathcal{S}$&MAE&$\mathcal{F}$&$\mathcal{S}$\\
		\midrule
		LEGS&{\tiny CVPR2015}&0.119&0.732&0.742&0.019&0.75&0.786&0.195&0.683&0.658&0.133&0.592&0.714&0.138&0.585&0.696\\
		MDF&{\tiny CVPR2015}&0.096&0.801&0.810&0.105&0.807&0.776&0.164&0.721&0.674&0.092&0.611&0.721&0.092&0.644&0.728\\
		ELD&{\tiny CVPR2016}&0.074&0.769&0.868&0.080&0.810&0.841&0.155&0.712&0.705&0.092&0.611&0.751&0.098&0.628&0.754\\
		DCL&{\tiny CVPR2016}&0.075&0.820&0.877&0.137&0.736&0.868&0.198&0.641&0.747&0.157&0.575&0.771&0.150&0.606&0.796\\
		KSR&{\tiny ECCV2016}&0.120&0.747&0。752&0.135&0.782&0.763&-&-&-&0.131&0.591&0.722&0.121&0.602&0.715\\
		RFCN&{\tiny ECCV2016}&0.089&0.835&0.859&0.107&0.834&0.852&0.169&{\color{blue} 0.743}&{\color{blue}0.794}&0.111&0.627&0.764&0.090&0.712&{\color{blue}0.859}\\
		DHS&{\tiny ECCV2016}&0.054&0.806&0.870&0.060&0.841&0.884&0.133&0.686&0.749&-&-&-&0.065&0.698&0.818\\
		NLDF&{\tiny CVPR2017}&0.048&0.838&0.879&0.063&0.839&0.875&0.130&0.708&{\color{red}0.889}&0.080&0.634&0.770&0.066&0.710&0.816\\
		Amulet&{\tiny ICCV2017}&0.052&0.813&0.886&0.059&0.841&0.894&0.140&{\color{green}0.755}&0.757&0.098&0.626&0.780&0.085&0.657&0.804\\
		SRM&{\tiny ICCV2017}&0.046&0.874&0.887&{\color{green}0.056}&{\color{blue}0.892}&0.895&0.132&0.671&0.741&0.069&{\color{blue}0.707}&0.798&0.059&0.678&0.836\\
		UCF&{\tiny ICCV2017}&0.062&0.823&0.875&0.069&0.852&0.883&0.169&0.644&0.753&0.120&0.628&0.760&0.117&0.588&0.782\\
		PAGRN&{\tiny CVPR2018}&0.048&{\color{blue}0.886}&0.887&0.064&0.891&0.889&-&-&-&0.072&{\color{green}0.711}&0.775&{\color{blue}0.055}&{\color{green}0.788}&0.838\\
		PICA&{\tiny CVPR2018}&{\color{green}0.042}&0.847&{\color{green}0.905}&{\color{red}0.047}&0.865&{\color{red}0.916}&{\color{red}0.108}&0.721&0.776&{\color{blue}0.068}&0.691&{\color{green}0.825}&{\color{green}0.054}&{\color{blue}0.748}&{\color{green}0.863}\\
		C2S&{\tiny ECCV2018}&0.046&0.848&{\color{blue}0.889}&{\color{blue}0.057}&0.860&{\color{blue}0.896}&{\color{blue}0.122}&0.702&0.760&0.072&0.698&0.799&0.062&0.686&0.831\\
		RA&{\tiny ECCV2018}&{\color{blue}0.045}&{\color{green}0.913}&0.887&0.059&{\color{green}0.896}&0.893&0.124&0.709&0.764&{\color{red}0.062}&0.701&{\color{blue}0.814}&0.059&0.723&0.839\\
		{\bfseries Ours}&~&{\color{red}0.041}&{\color{red}0.916}&{\color{red}0.909}&{\color{red}0.047}&{\color{red}0.916}&{\color{green}0.903}&{\color{green}0.114}&{\color{red}0.863}&{\color{green}0.815}&{\color{green}0.066}&{\color{red}0.743}&{\color{red}0.833}&{\color{red}0.047}&{\color{red}0.807}&{\color{red}0.884}\\
		\bottomrule
		
	\end{tabular}
	\label{MAEF}
\end{table*}
\subsection{Intractable Area F-measure Loss}\label{IAFloss}

 We observe that in multioutput encoder-decoder neural networks, outputs from different positions with different resolutions have different characteristics. Generally speaking, taking the deeply supervised multioutput network as an example, deeper outputs with low resolutions can capture semantic information better while shallower outputs with high resolutions concerns more on the spatial features. Some examples of outputs from different positions are shown in Fig. \ref{1}. As can be seen from Fig. \ref{1}, first, high-resolution output saliency maps are more precise than low-resolution maps at the boundary of objects: second, there are some interference objects that are easily misjudged and different outputs make different saliency judgments on them. Both the object boundary and interference objects are difficult points to improve the detection accuracy. The judgement ability in these areas is always a significant factor affecting the performance of a salient object detection model.
 \par Thus, we propose a new loss function to promote the model's performance in these areas. We need to find the intractable areas of images in the training set. First, we apply another dataset with fewer images to train the model for fewer iterations, and the training result is rough. Then, we test images in the training set utilizing the roughly trained model, and some saliency maps with different resolutions are obtained. For input image \emph{I}, there are five output saliency maps $S_i,i \in {\{1,2,3,4,5\}}$. We apply $S_1$  with the largest resolution and $S_5$ with the smallest resolution to calculate the difference map based on the observation that difference between high-resolution maps can only show the boundary area but fail to get the intractable area such as the disturbing objects. First, we bilinearly upsampled these two saliency maps to the resolution of the original image and obtain $S_{1}^{'}$ and $S_{5}^{'}$. Then the different areas can be obtained by the pixel-level comparison between $S_{1}^{'}$ and $S_{5}^{'}$ and the coordinate set \emph{C} of the different areas can be calculated as follows, for all coordinates $(i,j)$ in $S_{1}^{'}$ and $S_{5}^{'}$:
 \begin{equation}
 \left\{
 \begin{aligned}
 (i,j)\in C \quad {\rm if}\quad S_1^{'}(i,j)- S_5^{'}(i,j)=0\\
 (i,j)\notin C  \quad {\rm if}\quad S_1^{'}(i,j)- S_5^{'}(i,j)\ne0
 \end{aligned}
 \right
 .\end{equation}
 After getting the different maps, we train the model for the second time. For the second training, the saliency score is binarized, and the intractable area F-measure loss is calculated as follows:
 \begin{equation}
 L_f = 1-\frac{(1+\beta^2)\times P_c \times R_c}{\beta^2\times P_c+R_c}
 \end{equation}
 where $P_c$ and $R_c$ represent the precision and recall of area $C$. The formula of intractable area F-measure loss equals to 1 minus the F-measure of area $C$. The effectiveness can be understood from two aspects. On the one hand, the loss function is designed directly according to the evaluation metric, which is proved to be useful to promote the test results in a lot of computer vision tasks such as object detection \cite{64} and semantic segmentation \cite{63}; On the other hand, the IAF loss is only calculated on the intractable areas , thus promoting the model to process these areas more effectively and enhancing the generalization ability of the model in dealing with complex images.\par
 Note that, the second training is {\bfseries{not}} the fine-tuning of the model gained by the first training. The only purpose of the first training is to obtain the difference maps for the training set in the second training. When testing, only the model from the second training is applied to obtain saliency maps. Thus, our proposed method is end-to-end when testing though the training involves two processes.

\subsection{Training}\label{train}
Suppose that the multioutput neural networks can be divided into \emph{M} layers and that every layer of the decoder generates an output. Every output can produce a loss term. The final loss function can be defined as:
\begin{equation}
\begin{split}
L(I,G,{\rm W,w})=\beta l_f(I,G,{\rm W},{\rm w^{(1)}})\\+\sum_{m=1}^{{\emph M}}\alpha_m \emph l_{side}^{m}(I, G, \rm{W},\rm{w}^{({\emph M})})
\end{split}
\end{equation}
 where $\alpha_m$ is the weight of the cross-entropy loss in the $m^{th}$ layer and $\beta$ is the weight of intractable area F-measure loss. $I$ and $G$ represent the input image and its ground truth. Each output is obtained by a separate score function ${\rm w}^{(m)}$, and ${\rm w}$ refers to the set of all score fuctions:
 \begin{equation}
 {\rm w=(w^{1},w^{2},...,w^{\emph M})}
 \end{equation}\par
Here, $l_f(I,G,{\rm W},{\rm w^{(1)}})$ represents the intractable area F-measure loss function, and $l_{side}^{m}(I, G, \rm{W},\rm{w}^{({\emph M})})$ refers to the cross-entropy loss function of the $m^{th}$ output and can be calculated as follows:
\begin{equation}
\begin{split}
l_{side}^{m}(I, G, {\rm{W}},{\rm{w}}^{({\emph m})})= -\sum_{z=1}^{|I|}G(z)logP(G(z)=1|I(z),{\rm W},{\rm w^{\emph m}})&\\
-\sum_{z=1}^{|I|}(1-G(z))logP(G(z)=0|I(z),{\rm W},{\rm w^{\emph m}})&
\end{split}
\end{equation}\par
In the output-guided network, $M$ equals 5 so that 5 outputs are gained. Instead of fusing these outputs as in\cite{17} by adding additional computing, we directly apply the output of the first layer, which has the highest resolution, as our final saliency score. Considering that the output of the first layer has the highest importance, $\alpha_1$ is set higher than others, the weights of all the loss functions are:
\begin{equation}
\{\alpha_1, \alpha_2, \alpha_3, \alpha_4, \alpha_5, \beta\}=\{50,4,4,4,4,25\}
\end{equation}

\begin{figure*}
	\centering
	\includegraphics[width=0.92\linewidth]{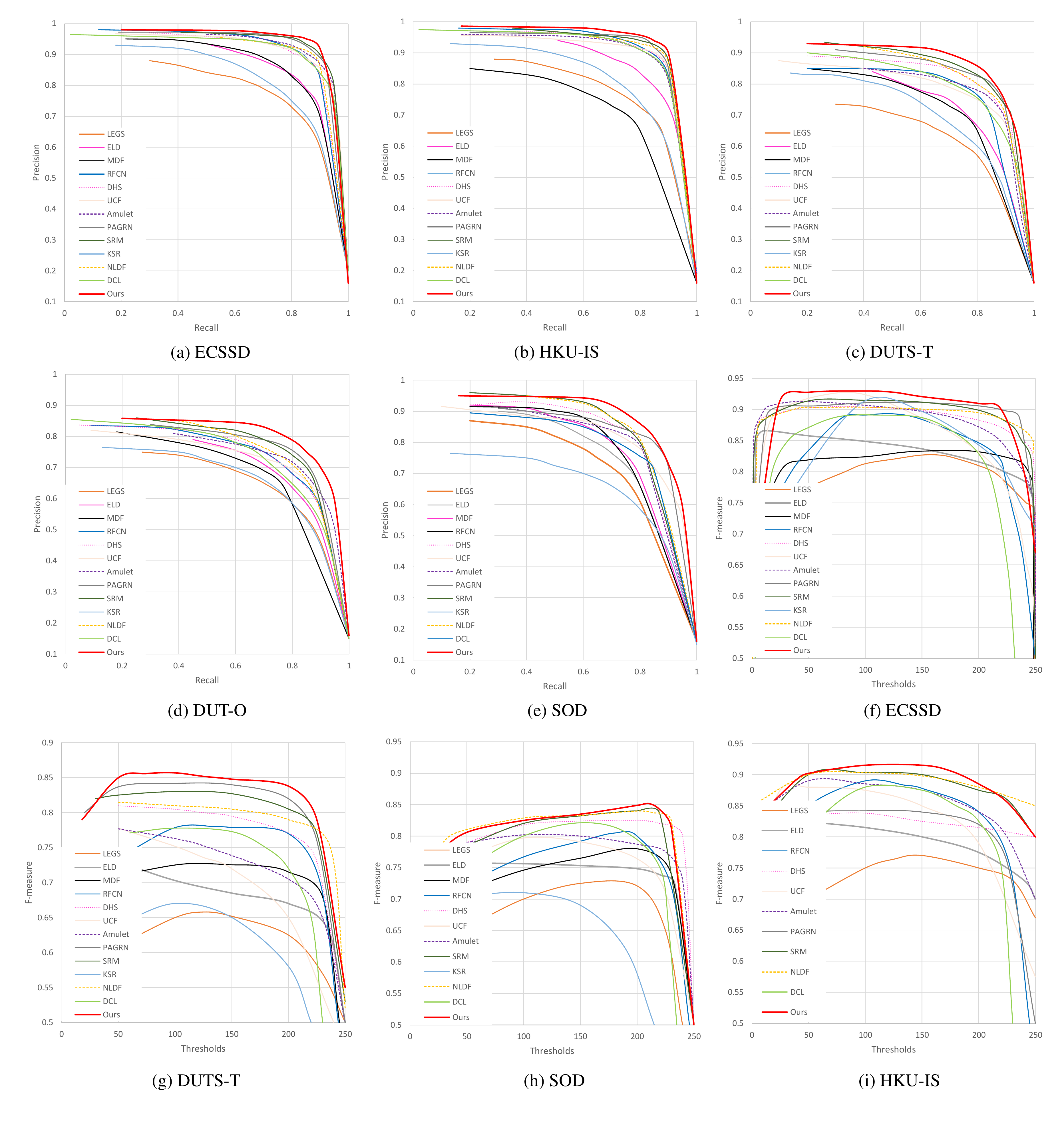}
	\caption{(a)-(e) are P-R curves on various datasets, including ECSSD, HKU-IS, DUTS-T, DUT-O and SOD. (f)-(i) are F-measure curves on various datasets, including ECSSD, DUTS-T, SOD and HKU-IS.}
	\label{FPR}
\end{figure*}
\section{Experimental Results} \label{exp}
\subsection{Implementation Details}
  We use the PyTorch  framework to train and test our model. All images are resized to $320\times 320$ pixels for training and testing. We select SGD with a weight decay of 0.0005 and a momentum of 0.9 as the optimizer. Inspired by\cite{40}, we use the `poly' policy to set the learning rate. For an iteration, its learning rate equals the initial learning rate multiplied by $(1-\frac{iter}{maxiter})^{power}$, where the initial learning rate is set to 0.0001 and power is set to 0.9. Due to the use of IAF loss, the model needs to go through two separate training processes, for which we used the same parameter configurations. It takes approximately 21 hours to train 40 epochs on a sever with an NVIDIA Titan X GPU (with 12G memory).
   \begin{figure*}
  	\centering
  	\includegraphics[width=1\linewidth]{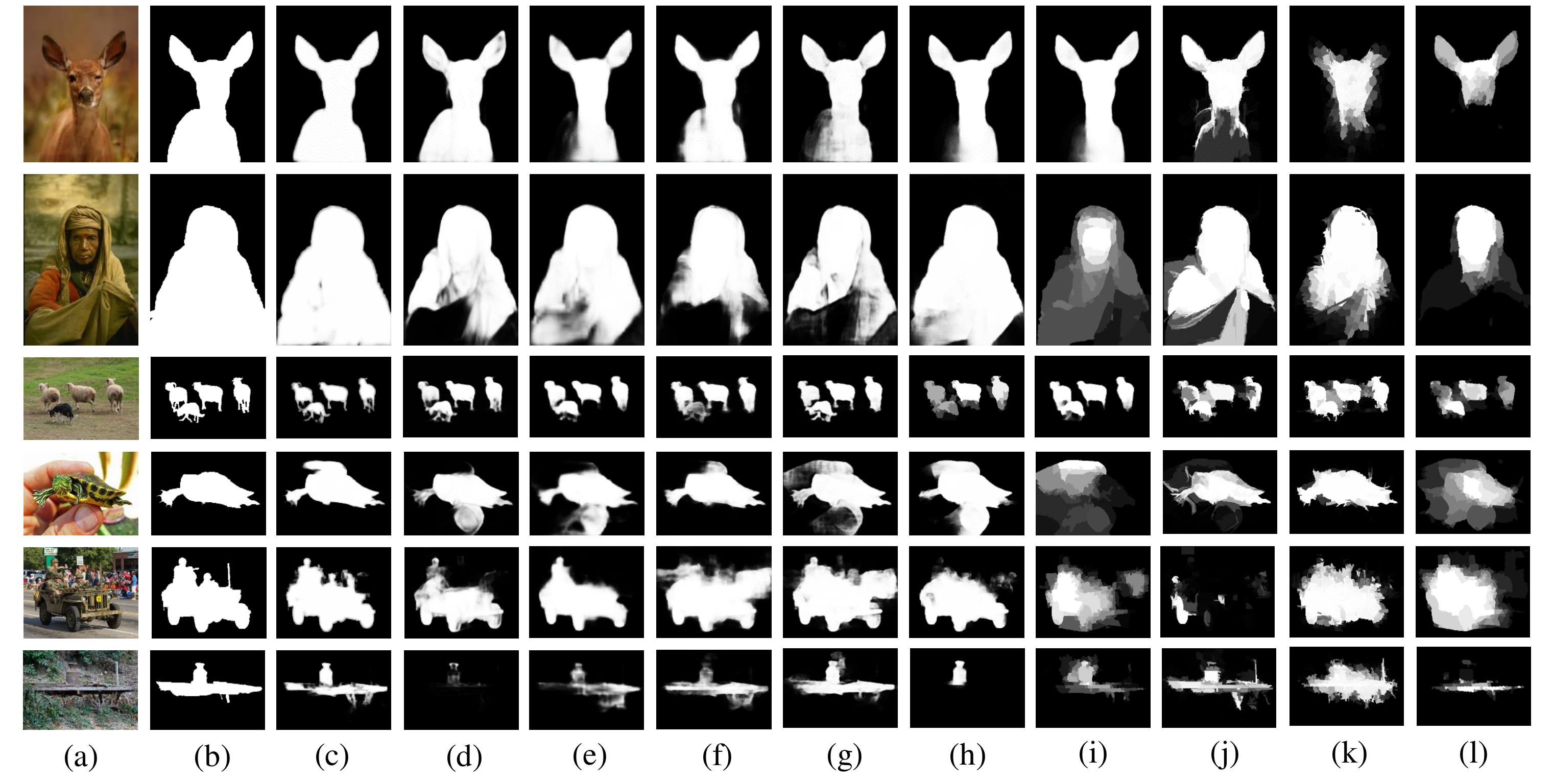}
  	\caption{Visual comparison with 9 state-of-the-art methods. (a) Input image; (b) ground truth; (c) ours; (d) PAGRN\cite{35}; (e) SRM\cite{29}; (f) Amulet\cite{28}; (g) UCF\cite{26}; (h) NLDF\cite{44}; (i) KSR\cite{48}; (j) MDF\cite{24}; (k) ELD\cite{46}; (l) LEGS\cite{23}. Our method performs best for images with various characteristics.}
  	\label{visual}
  \end{figure*} 
 \subsection{Datasets and Evaluation Metrics}
 \subsubsection{Datasets}
Six datasets are used to train and test our models: MSRB\cite{45}, DUTS\cite{41}, ECSSD\cite{7}, DUT-OMRON\cite{42}, HKU-IS\cite{24}, and SOD\cite{43}.\\
{\bfseries MSRB}: This dataset contains 5000 high quality images with high precision marks. These images are abundant in species, but their backgrounds are usually simple.\\
{\bfseries DUTS}: This dataset includes 10553 images for training and 5019 images for testing. This dataset’s images are characterized by a large quantity of abundant species and high marked quality.\\
{\bfseries ECSSD}: This dataset contains 1000 images with a complex background, and the ground truth of the image in the dataset usually contains very rich semantic information.\\
{\bfseries DUT-OMRON}: This dataset contains 5168 high quality images. The images of this dataset include one or more salient objects and their backgrounds are very complicated. It is relatively more difficult to achieve salient object detection on these images. Hence, it is a significant dataset to determine whether a salient object detection model can perform well for complex images.\\
{\bfseries HKU-IS}: This dataset contains 4447 high-precision labeled images. Images in the dataset are often equipped with many salient objects, and some of these salient objects are located at the edge of the images, which brings a great challenge to salient object detection.\\
{\bfseries SOD}: This dataset contains 300 images. These images' background and the shape of the salient objects are quite complex. It is a very challenging dataset. \par
 We use MSRB to train our model for the first time and then use this model to test the training set of DUTS and obtain the difference maps. Then, the training set of DUTS and the difference maps are used for the second training to obtain the final model. The test set of DUTS and other datasets are used to test the model.

\subsubsection{Evaluation Metrics}
We utilize four methods that are extensively applied in the salient object detection field to test our model’s performance on test sets: precision-recall (PR) curves, F-measure and mean absolute error (MAE) and S-measure. The saliency maps are binarized by varying the threshold from 0 to 255, and pairs of precision and recall under different thresholds are computed to plot the PR curve. Then, the saliency map is binarized with a fixed threshold, which is determined as twice the mean saliency value of the saliency map. The F-measure is calculated as follows:\\  
\begin{equation}
F_\beta=\frac{(1+\beta^2) \times Precision\times Recall}{\beta^2 \times Precision + Recall}
\end{equation}
Similar to most other methods\cite{26,28,29}, we set $\beta^2$ to 0.3, making the precision's influence factors larger than that of the recall.\par 
Due to the binarization of the saliency map, the F-measure cannot directly measure the difference between the ground truth and the saliency map obtained by the model. Hence, we also apply MAE, which values the average pixelwise absolute difference between the saliency map and binary ground truth:\\
\begin{equation}
{\rm MAE}=\frac{1}{W\times H}\sum_{x=1}^{W}\sum_{y=1}^{H}|S(x,y)-G(x,y)|
\end{equation}
where W and H are the width and height of the saliency map $S$, respectively.\par
Structure measure (S-measure) \cite{65} is a new evaluation metric to evaluate region-aware and object-aware structural similarity between saliency maps and ground truth maps. It can be calculated as follows:
\begin{equation}
S=\alpha * S_{o}+(1-\alpha) * S_{r}
\end{equation}
where $S_{o}$ and $S_{r}$ represent object-aware and region-aware structural similarity, respectively. $\alpha$ is set to 0.5. A model with a higher F-measure score ,lower MAE score and higher S-measure score has better performance.

 \subsection{Performance Comparison}
  We compare 15 state-of-the-art classic salient object detection methods, including LEGS\cite{23}, ELD\cite{46}, MDF\cite{24}, KSR\cite{48}, DCL\cite{49}, RFCN\cite{47}, NLDF\cite{44}, DHS\cite{18}, UCF\cite{26}, Amulet\cite{28}, PAGRN\cite{35}, SRM\cite{29}, C2S\cite{53}, RA\cite{21} and PICA\cite{36}. Most of these methods are based on deep learning.
   
  \subsubsection{Qualitative Evaluation}
    Fig. \ref{visual} displays the visual comparison between our method and the others. Our method can judge the salient object better and more accurately display the area of the salient object. Our method performs much better than the other methods in the following challenging situations:\par
  (1) When confronting multiple salient objects in an image, our method makes more accurate decisions on multiple salient objects. As shown in the third line, our method precisely judges all four salient objects.\par
  (2) When the shape of the salient object is complicated, our algorithm still demonstrates the shape of the salient object obviously and favorably. As shown in the fifth line, although the salient object's upper edge contour is quite complex and the rough sketch feature is quite blurry, our method precisely recovers the rough sketch of the salient object and do not generate an erroneous judgement.\par
  (3) Thanks to the introduction of the attention mechanism, when salient objects are surrounded by some interferential factors disturbing the salient judgement, our method is able to perform better and had strong antijamming ability. For example, in the first line, the lower left quarter of the salient object is highly similar to the surrounding areas, so it is quite easy to cause an erroneous judgement. Our method makes a very accurate judgement, while most of the other state-of-the-art methods incorrectly judge that area as nonsalient.

 \subsubsection{Quantitative Evaluation}
 The PR curves and F-measure curves are shown in Fig. \ref{FPR}. For a PR curve, a higher precision and slower attenuation represents a better performance. Compared with the other methods, our method has the best performance on all the datasets.\par 
In Table \ref{MAEF}, we also compare our method with the state-of-the-art methods in terms of MAE, F-measure and S-measure. For MAE score, we obtained the best performance on most of the datasets. Although we did not realize the best performance on SOD and DUT-OMRON on MAE, our method demonstrates high competition. For the F-measure score, our method performs the best on all the datasets. Compared with the second-ranked method, our method improves the F-measures score by  4.8\%, 2.7\%, 14.3\%, 4.5\% and 2.4\% on HKU-IS, ECSSD, SOD, DUT-OMRON and DUTS-TE, respectively. For S-measure score, our method performs best on three datasets and and ranks second on another two datasets.\par
Based on the indexes being synthesized, in comparison to the other state-of-the-art methods, our method shows the best performance overall. The excellent execution on all datasets demonstrates that our method possesses stronger universality.

 \begin{table}[t]
 	\centering
 	\caption{Ablation experiments on DUTS and SOD.}
 	\begin{tabular}{clcc}
 		\toprule
 		No.&settings&SOD&DUTS\\
 		\midrule
 		(a)&Comparison of attention module\\
 		1&basline&0.13082&0.05323\\
 		2&+SE&0.12433&0.05232\\
 		3&+CBAM&0.12234&0.05185\\
 		4&+OGAM&\textbf{0.11619}&\textbf{0.04895}\\
 		\midrule
 		(b)&Comparison of IAF loss\\
 		5&baseline(No.4 setting)&0.11619&0.04895\\
 		6&BCE loss+IAF loss&\textbf{0.11362}&\textbf{0.04658}\\
 		\midrule
 		(c)&Comparison of residual blocks\\
 		7&baseline(No.4 setting)&\textbf{0.11619}&\textbf{0.04895}\\
 		8&without residual block&0.11794&0.04978\\
 		\bottomrule

 	\end{tabular}
 \label{multi}
 \end{table}
\subsubsection{Memory Comparison}
\begin{figure}
  \centering
  \includegraphics[width=1\linewidth]{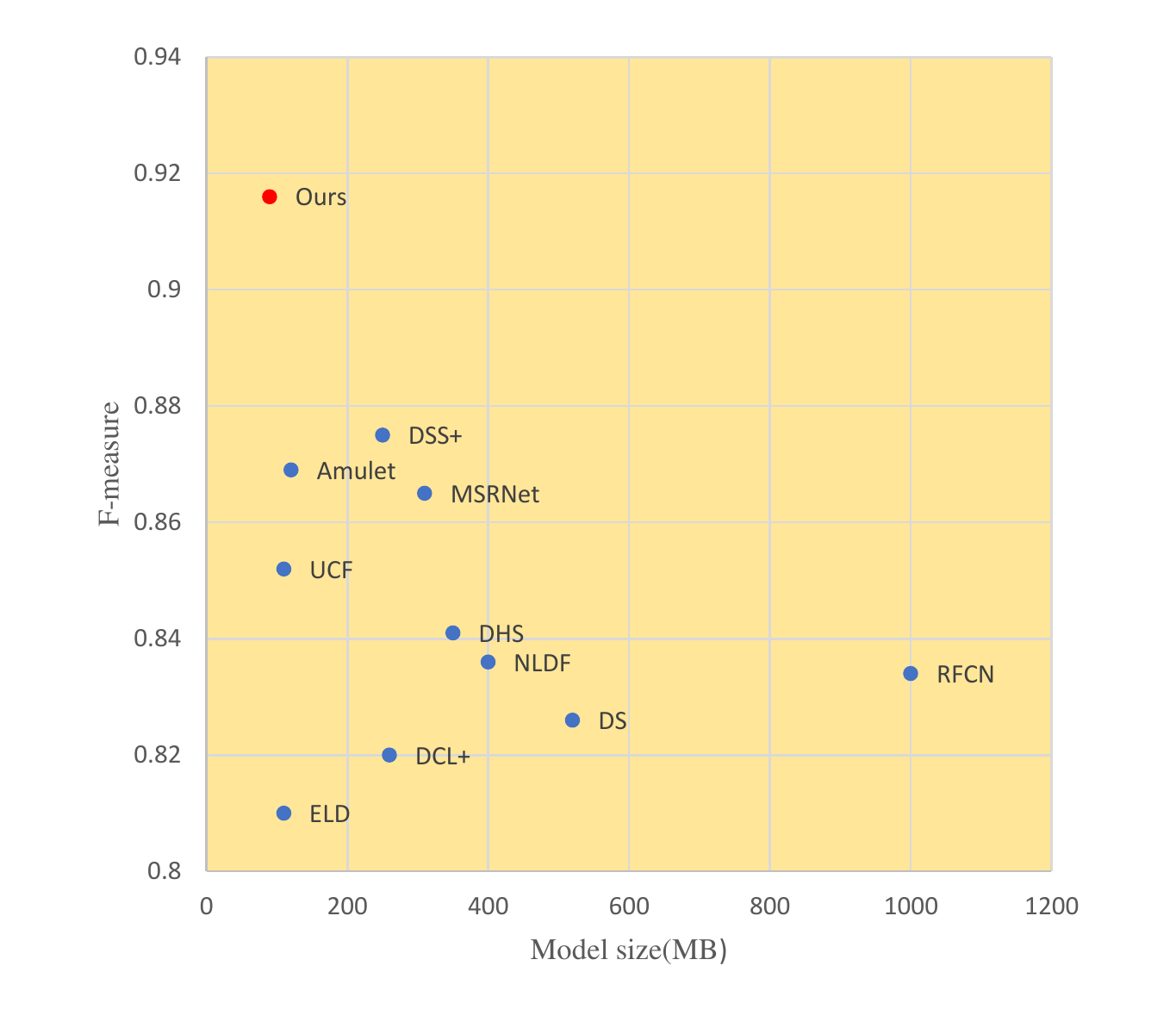}
  \caption{Memory comparison with some methods, including Amulet[], DSS+\cite{17}, MSRNet\cite{55}, UCF\cite{26}, DHS\cite{18}, NLDF\cite{44}, RFCN\cite{47}, DS\cite{56}, DCL+\cite{49} and ELD\cite{46}.}
  \label{mem}
  \end{figure}

 The algorithms based on deep learning usually require a large computation and memory footprint. In general, a deeper neural network can gain better performance, but it is also followed by a larger memory footprint and computation so it is difficult to apply the model to real-time detection and to use it on mobile terminals, which reduces the practicability. Hence, the size of the neural network model is also one of the significant factors when measuring a salient object detection algorithm based on deep learning. Fig. \ref{mem} shows some methods' model size and F-measure on ECSSD. The model size of many methods is very large, while those with smaller model sizes usually have a general effect. Our method is the only one with a model size of less than 100 MB and an F-measure score higher than 0.9. Our model is lightweight but very effective.
 
 \subsection{Ablation Studies}
 
 \subsubsection{Evaluation of output-guided attention}
 As shown in table \ref{multi}, to verify the effect of the output-guided attention module, we compare the effects of models with and without the output-guided attention module. The experimental results show that the output-guided attention module can greatly improve the model's effect. In addition, we also test the effects of some other types of attention modules on model improvement. Two attention modules are tested: SE\cite{32} and CBAM\cite{34}. Different from the output-guided attention module, SE only uses channel attention, and both SE and CBAM only take the processed feature maps themselves as input. The experimental results show that the effect of using SE alone is not significant enough and CBAM utilizing both channel attention and spatial attention can produce better effects. Our output-guided attention module performs best among these three kinds of attention modules.
  \begin{figure}[t]
 	\centering
 	\includegraphics[width=1\linewidth]{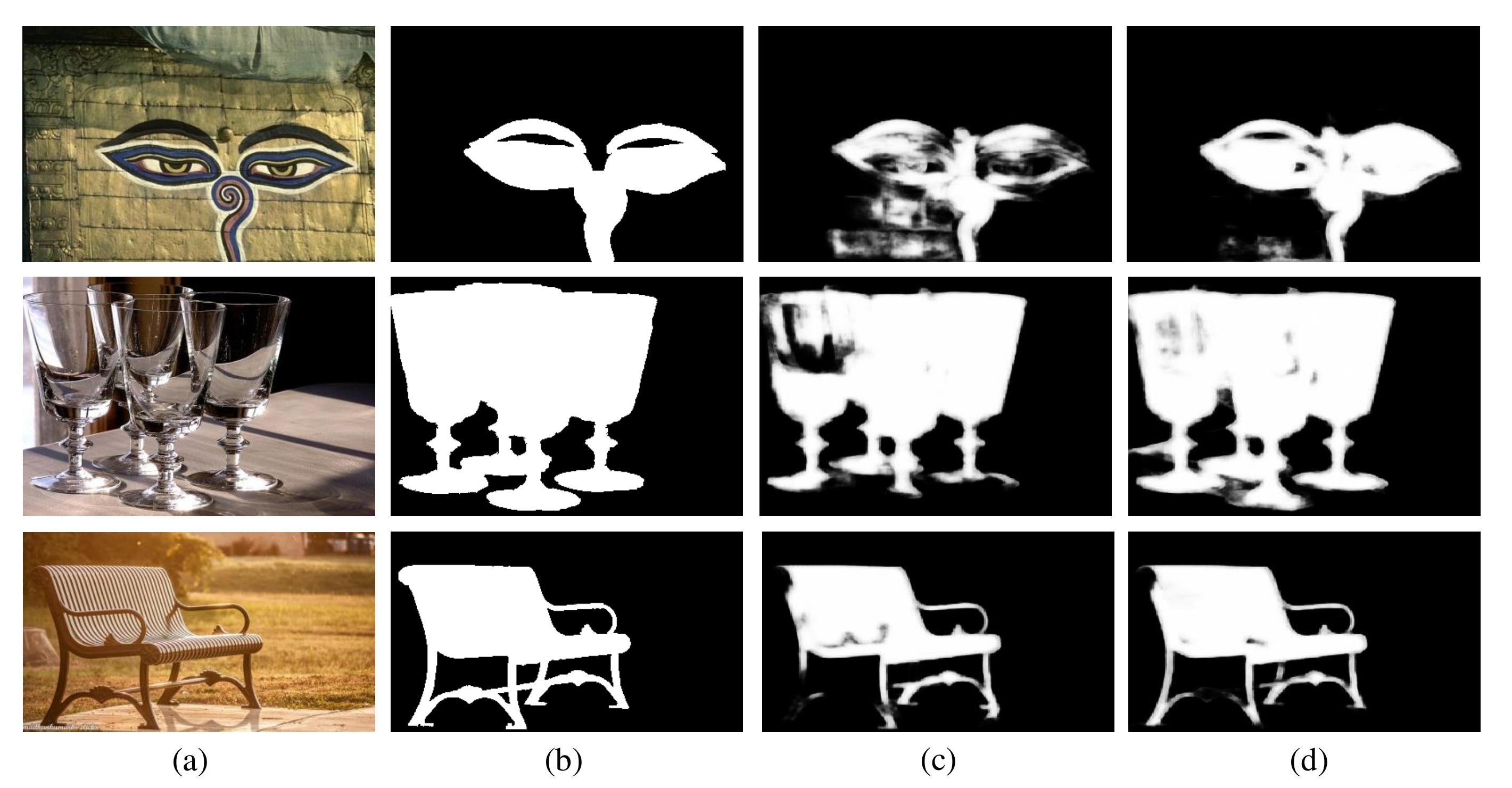}
 	\caption{Comparison between output obtained by models applying IAF loss and not applying IAF loss. (a) Input image; (b) ground truth; (c) output of model not applying IAF loss; (d) output of model applying IAF loss. }
 	\label{loss}
 \end{figure} 
 \subsubsection{Evaluation of intractable area F-measure loss}
  The intractable area F-measure loss is used to upgrade the model's judgement ability when encountering difficult areas. To test its effectiveness, we test the performance of models trained applying the intractable area F-measure loss and not applying the intractable F-measure loss on SOD and DUTS-TE, respectively. As shown in Fig. \ref{loss}, comparing the test results of the two models, the model performs better in the marginal areas of the salient objects and makes a more precise judgment of the difficult areas after utilizing the intractable area F-measure loss. Quantitative analysis is shown in Table \ref{multi}. After utilizing the intractable F-measure loss, the MAE score on both datasets decline.
  \subsubsection{Ablation of residual block}
  Residual blocks can make a very deep neural network easier to train and improve the effect of the neural network. To test the residual blocks' influence on our model, we eliminate the original residual block of the model and then test its performance. The experimental results are shown in Table \ref{multi}. Observing the training process, we find that the model with the residual blocks converged faster and that the final loss value was smaller. The application of the residual blocks slightly raised the model's effects. Thus, we deemed that the utilization of residual blocks in our model causes overfitting. 	
  \subsubsection{Selection of convolution size}
\begin{figure}
  \centering
  \includegraphics[width=1\linewidth]{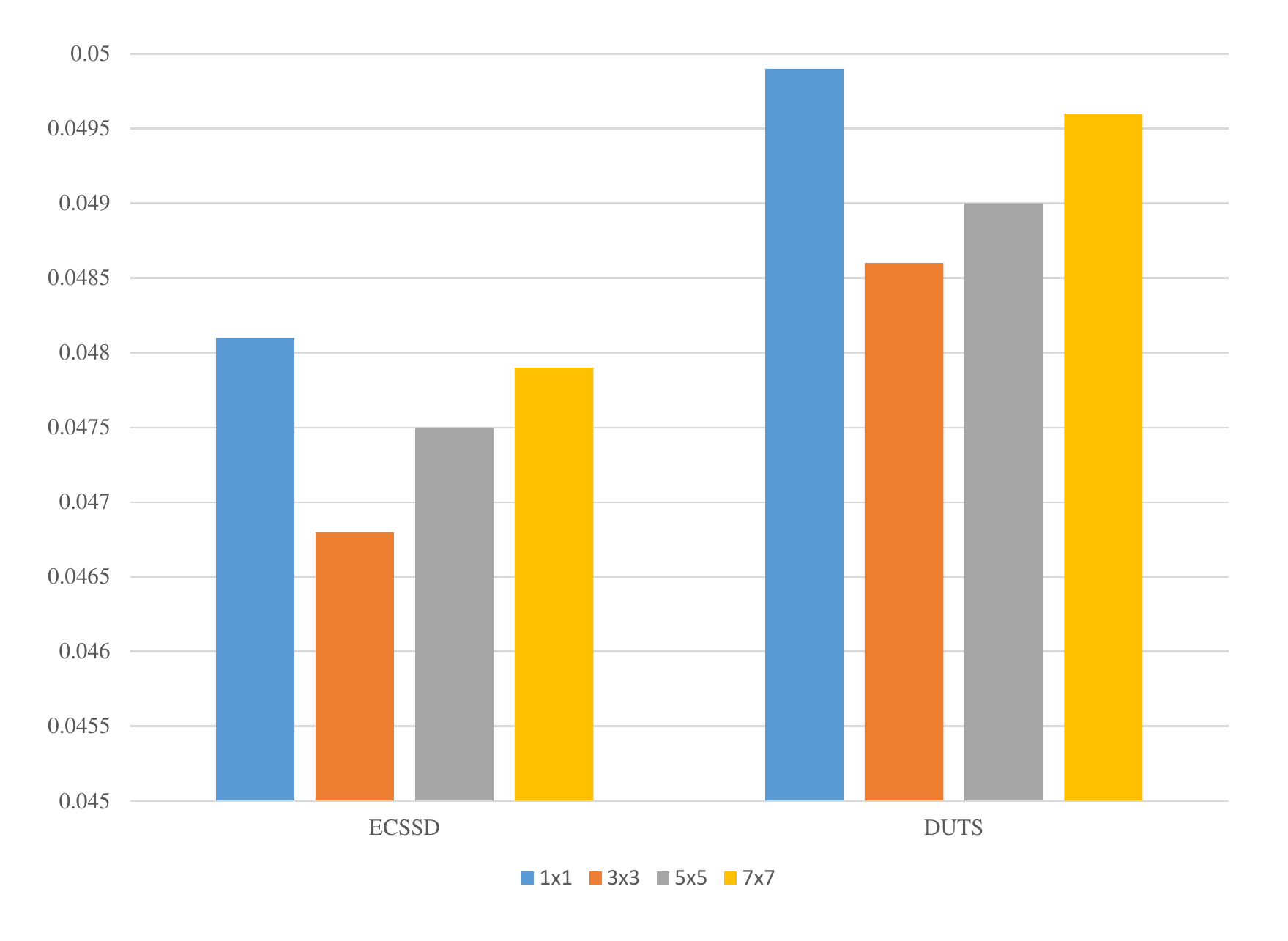}
  \caption{Comparison of MAE score of four sizes of convolution on ECSSD and DUTS. }
  \label{size}
  \end{figure}

The choice of convolution size has a great influence on the performance of convolutional neural networks. DSS\cite{17} uses a large convolution to process the feature maps extracted from the encoder in every layer. Theoretically, a large convolution can increase the receptive field and extract more semantic information, so it is used to process feature maps from encoders that do not sufficiently extract semantic information compared with those from decoders. Inspired by DSS, we first choose a convolution of size $7\times 7$ but find that the performance of the model unexpectedly became worse. To determine the most suitable convolution size, we test the convolution of four sizes: $7\times 7$, $5\times 5$, $3\times 3$, $1\times 1$. The performance of these models on five datasets are shown in Fig. \ref{size}. The lowest MAE score on all five datasets is achieved by the model with a $3\times 3$ convoluton. The receptive field of a $1\times 1$ convolution is too small to integrate the information extracted from the encoder and overfitting is caused by a $5\times 5$ convolution and a $7\times 7$ convolution, which are too large. Feature maps extracted from encoders are mainly used to better restore the shape of salient objects, so spatial information is more important than semantic information. A large convolution may destroy the spatial information, which is harmful for the accurate display of salient objects.
  
  \subsubsection{Application of output-guided attention in the other models} 
\begin{figure}
\centering
\includegraphics[width=1\linewidth]{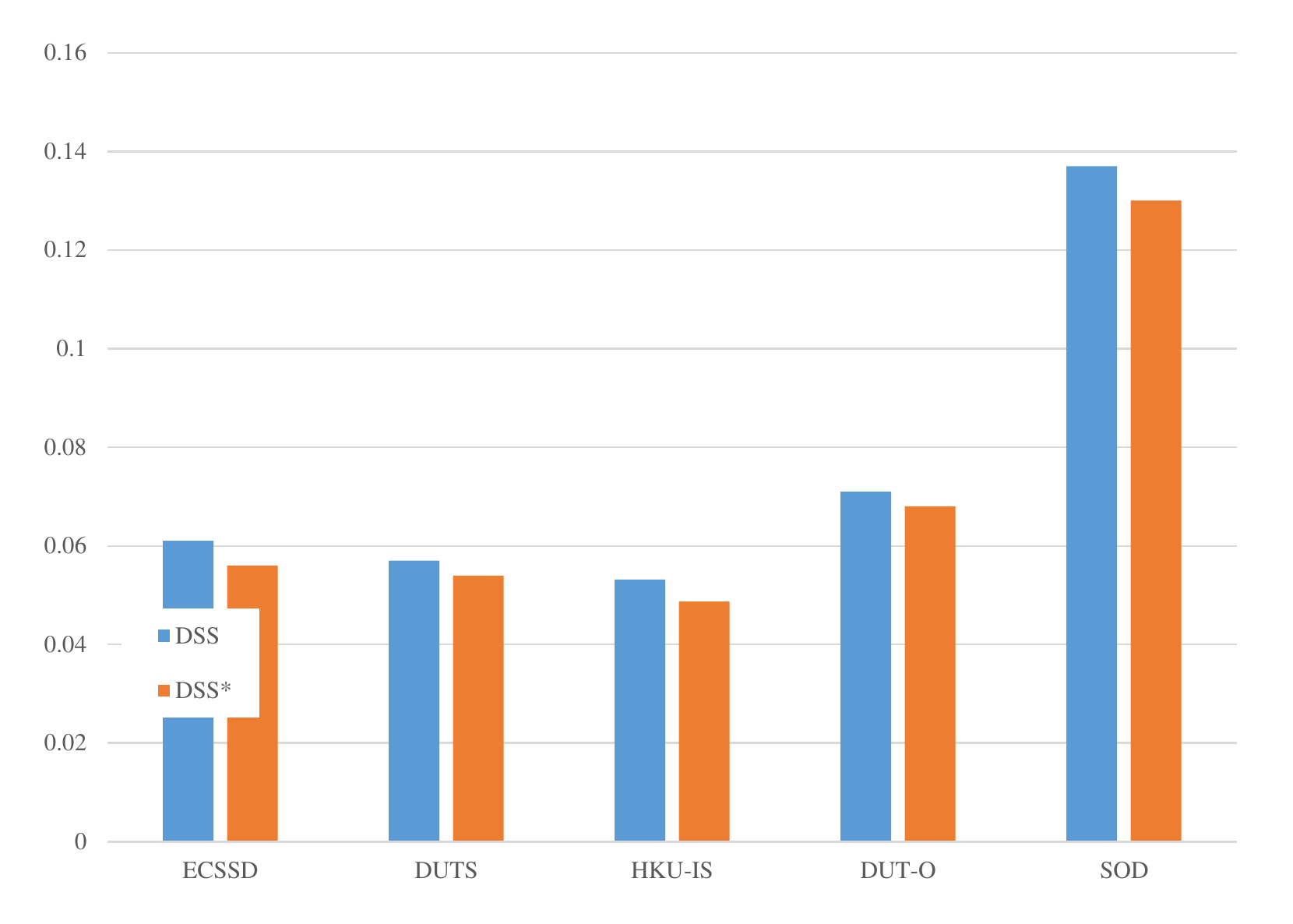}
\caption{Comparison of MAE scores on five datasets of the original DSS and DSS*(DSS applying output-guided attention module and IAF loss.)}
\label{DSS}
\end{figure}

The output-guided attention module proposed by us in this paper is a lightweight and universal module that can be used in all multioutput models. We test the effect of the output-guided attention module on some other multioutput models. DSS\cite{17} is a classic salient object detection model with multiple outputs. The original DSS uses two convolutional layers to process each side output, and we add an output-guided attention module after the first convolutional layer. The experimental results are shown in Fig. \ref{DSS}, where the MAE score of the five datasets between the original DSS and the DSS with the output-guided attention module are compared. Compared with the original model, the MAE score of the five datasets after using the output-guided attention module decreases by 8.9\%, 4.6\%, 8.0\%, 4.2\%, and 5.1\%, respectively.

\section{Conclusion}\label{conclusion}
 In this paper, we proposed a new output-guided attention module. Experimental results show that compared with other attention modules, the output-guided attention module constructed by the processed feature maps themselves and other resolution outputs can reduce errors and achieve better performance. Our proposed model, based on output-guided attention, showed outstanding performance on multiple datasets. Owing to the output-guided attention module, our model has stronger robustness. The proposed intractable area F-measure loss can effectively improve the performance of the model when facing images with complex backgrounds and salient objects with complicated shapes. The improvements of the output-guided attention module and intractable area F-measure loss on other multioutput methods demonstrate that these two methods are universal. We suggest that researchers try to use the output-guided attention module and intractable area F-measure loss when constructing other neural networks for salient object detection. We believe that ‘blind overconfidence’ is a common problem faced by many attention modules in salient object detection and that the output-guided attention module provides a new way to solve this problem. In the future, we will further explore additional ways to solve the problem of `blind overconfidence'.
 
\bibliographystyle{ieeetr}
\bibliography{1}

\begin{IEEEbiography}[{\includegraphics[width=1in,height=1.25in,clip,keepaspectratio]{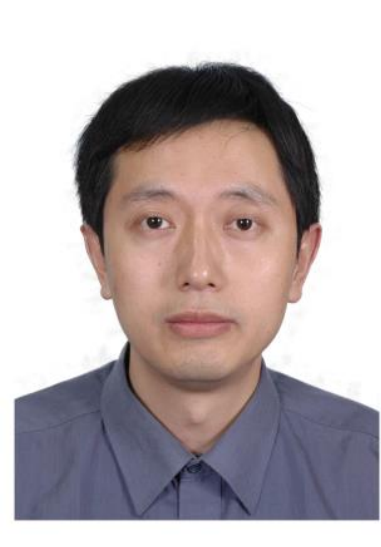}}] 
{Shiping Zhu} (M’05) received the B.Sc. and M.Sc. degrees in measuring and testing technologies and instruments from Xi’an University of Technology, Xi’an, China, in 1991 and 1994, respectively, and the Ph.D. degree in precision instrument and machinery from Harbin Institute of Technology, Harbin, China, in 1997.\par
From 1997 to 1999, he was a Postdoctoral Fellow with Beihang University, Beijing, China. From 2000 to 2002, he was a Postdoctoral Fellow with the Brain and Cognition Research Center, Université Paul Sabatier, Toulouse, France. From 2002 to 2004, he was a Postdoctoral Fellow with the Department of Computer Science and Department of Electrical and Computer Engineering, Université de Sherbrooke, Sherbrooke, QC, Canada. Since 2005, he has been an Associate Professor with the Department of Measurement Control and Information Technology, School of Instrumentation Science and Optoelectronics Engineering, Beihang University. He has authored or coauthored more than 80 journal and conference papers. He received the second prize of National Technological Invention Award in 2013. He is the holder of 50 China invention patents. His current research interests include image processing and video coding, stereo matching, saliency detection and image/video object segmentation.
\end{IEEEbiography}
\begin{IEEEbiography}[{\includegraphics[width=1in,height=1.25in,clip,keepaspectratio]{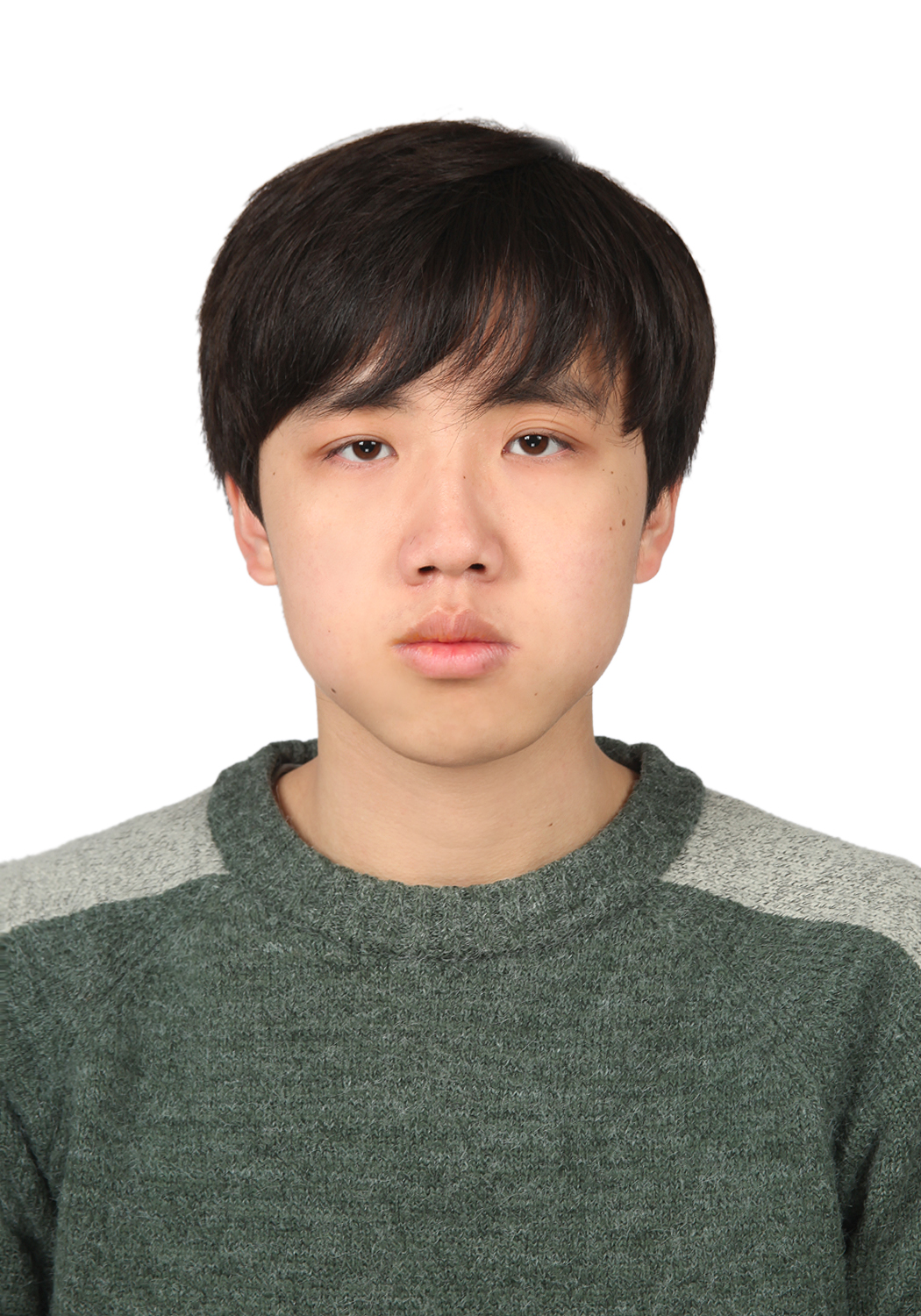}}] 
	{Lanyun Zhu} is currently pursuing the B.Sc. degree with Beihang University, Beijing, China. He is a research assistant with school of instrumentation and optoelectronic engineering, Beihang University. His currently research interests manly focus on computer vision, image processing and deep learning. 
\end{IEEEbiography}
\end{document}